\documentclass{article}

\usepackage[english]{babel}

\usepackage[letterpaper,top=2cm,bottom=2cm,left=3cm,right=3cm,marginparwidth=1.75cm]{geometry}

\usepackage{amsmath}
\usepackage{graphicx}
\usepackage[colorlinks=true, allcolors=blue]{hyperref}

\usepackage{epstopdf}
\usepackage{lipsum}

\title{Data-Driven Deepfake Image Detection Method - The 2024 Global Deepfake Image Detection Challenge}
\author{Xiaoya Zhu, Yibing Nan, Shiguo Lian\\AI Innovation Center, China Unicom}
\date{October 23, 2024}

\begin{document}
\maketitle

\begin{abstract}
With the rapid development of technology in the field of AI, deepfake technology has emerged as a double-edged sword. It has not only created a large amount of AI-generated content but also posed unprecedented challenges to digital security. The task of the competition is to determine whether a face image is a Deepfake image and output its probability score of being a Deepfake image. In the image track competition, our approach is based on the Swin Transformer V2-B classification network. And online data augmentation and offline sample generation methods are employed to enrich the diversity of training samples and increase the generalization ability of the model. Finally, we got the award of excellence in Deepfake image detection.
\end{abstract}

\section{Introduction}
As artificial intelligence advances swiftly, deepfake technology has arisen, embodying both promise and peril. It has not only given rise to a multitude of AI-crafted materials but also introduced hitherto unforeseen digital security threats. The Inclusion Global Multimedia Deepfake Challenge aims to invite participants to develop, test, and further evolve more accurate, effective, and innovative detection models against various types of Deepfake attacks in real-world scenarios, as well as to motivate innovative defense strategies and improve Deepfake recognition accuracy. 

\noindent The released MultiFF data set is built based on the "Technical Specifications for Financial Applications of Fake Digital Face Detection" launched by Ant Group. It is deeply combined with the actual defensive experience in the digital business to include truly diversified deep forgery attacks, which mainly contain: Wide range of attack types: including face-swapping, activation, attribute editing, full-face synthesis, audio-driving face generation, face restoration, digital adversarial samples, automated PS, and other common types; Multiple generation methods: More than 50 Deepfake generation methods are applied to the entire data set. In addition to the classic GANs model, the generation model also fully covers the new generative AI model of the Diffusion series. The attack generation paradigm includes image-to-image generation, text-to-image generation, text-to-speech generation, text-to-video generation, etc. The MultiFF data set released in this competition is much more comprehensive and diversified in multiple dimensions such as types, generation methods, generation paradigms, and face material attributes.

\noindent The task of the competition is to determine whether a face image is a Deepfake image and output its probability score of being a Deepfake image. Our approach is based on the Swin Transformer V2-B classification network. And online data augmentation and offline sample generation methods are employed to enrich the diversity of training samples and increase the generalization ability of the model. Swin Transformer addresses the discrepancy in adapting Transformer from the linguistic to the visual side by shifting window computational features to limit self-attentive computation to non-overlapping localized windows while also allowing crosswindow connections to improve efficiency. The hierarchical architecture has the flexibility to model at a variety of scales and has linear computational complexity with respect to image size. These features of Swin Transformer make it compatible with a wide range of visual tasks, including image classification and dense prediction tasks. Meanwhile, diversified training samples are generated for a wide variety of deepfake methods, including random facial region Cutout, local cropping, cartoonization, etc. Finally, for the classification prediction results, post-processing is performed by combining the Dlib third-party library face keypoint detector with the Opencv Haar cascade face detector to correct the confidence level of the network that is difficult to make a judgment.

\begin{figure}
\centering
\includegraphics[width=1\textwidth]{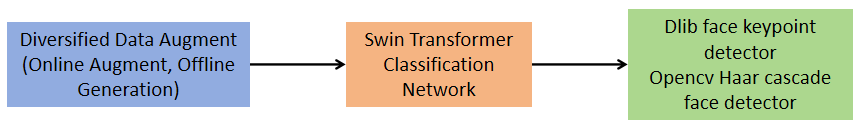}
\caption{\label{fig:figure1}Overview of our method.}
\end{figure}

\section{Method}

\subsection{Data preprocessing techniques applied}

 For a given input image, after resizing it to 256×256, data augment is performed in the following order to increase data diversity and generalization of the model.

\begin{enumerate}
\item Random horizontal flip with a probability of 0.5;
\item AutoAugment: the data augmentation method used in Swin Transformer is adopted, which mainly includes random image inversion, random contrast adjustment, random rotation, etc. (parameters kept the same as the settings in Swin Transformer).
\end{enumerate}

\noindent At the same time, some of the existing training data are transformed in order to generate more diverse training negative samples. The methods mainly include:

\begin{enumerate}
\item Random facial region cutout: 68 keypoints in the face image are detected using the dlib third-party library and it is randomly erased and filled with random colors with a probability of 0.5 for regions such as eyebrows, eyes, nose, lips, and so on;
\item Local cropping: a randomly selected 150×150 area of the image is enlarged to the original image size of 512×512 as a negative sample;
\item Random grayscaling, translating and overlaying: after grayscaling an image, translate it 4 times in a random direction and overlay the results of the 4 translations to form a new image;
\item Cartoonization: the image is first converted to a grayscale map, and the edges are detected using adaptive thresholding, then the number of colors is reduced by K-means clustering. The median filter is applied to the image to smooth, and finally the smoothed image is combined with the edge maps, thus generating a cartoon style image;
\item Sketch: After grayscaling the image, invert the grayscale image, fuzzy invert the image, invert the fuzzy image, and calculate the ratio of the original grayscale image to the inverted fuzzy image to get the effect of pencil sketching;
\item Binarization: randomly select a threshold near 128, set pixels above the threshold to 255 (white) and pixel points below the threshold to 0 (black).
\end{enumerate}

\noindent By employing specific data augment techniques, a series of diverse training samples are generated to withstand various deepfake methods(as shown in Fig.2), thus effectively improving the generalization performance and robustness of the model.

\begin{figure}[!htbp]
	\centering
	\begin{minipage}{0.3\linewidth}
		\centering
		\includegraphics[width=0.9\linewidth]{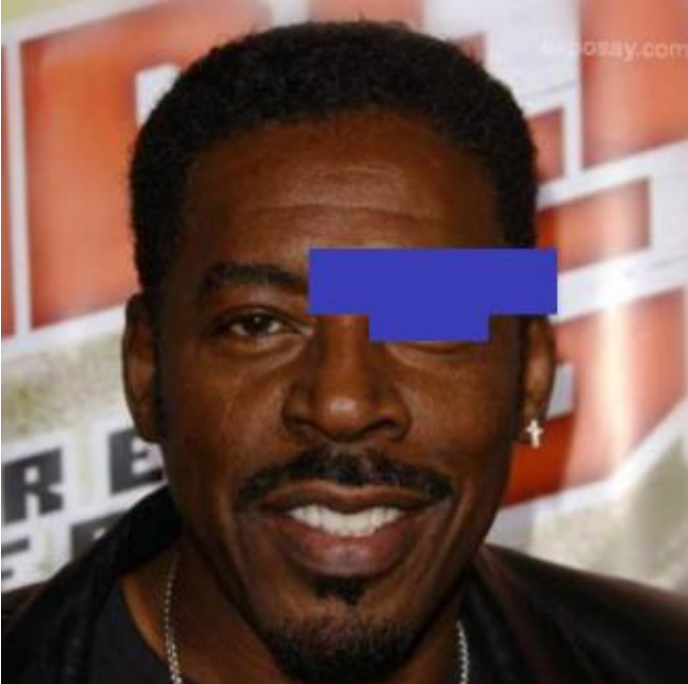}
	\end{minipage}
	\begin{minipage}{0.3\linewidth}
		\centering
		\includegraphics[width=0.9\linewidth]{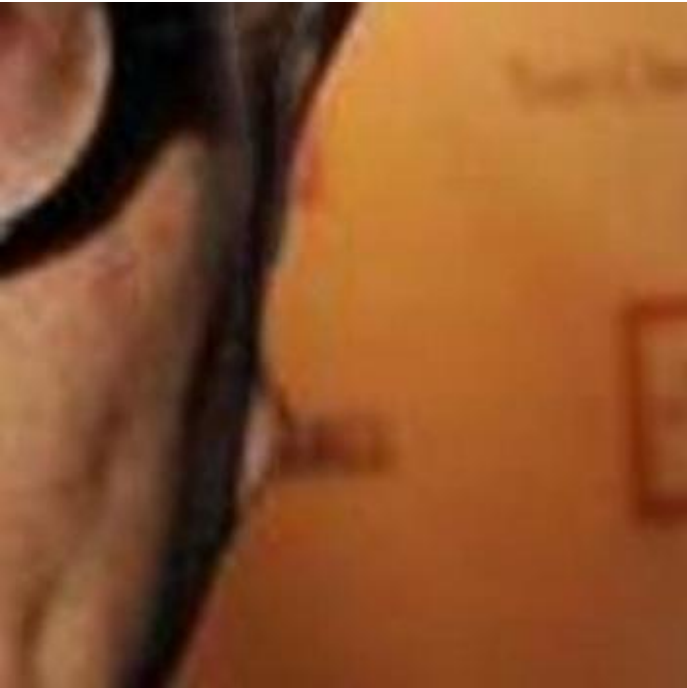}
	\end{minipage}
    \begin{minipage}{0.3\linewidth}
		\centering
		\includegraphics[width=0.9\linewidth]{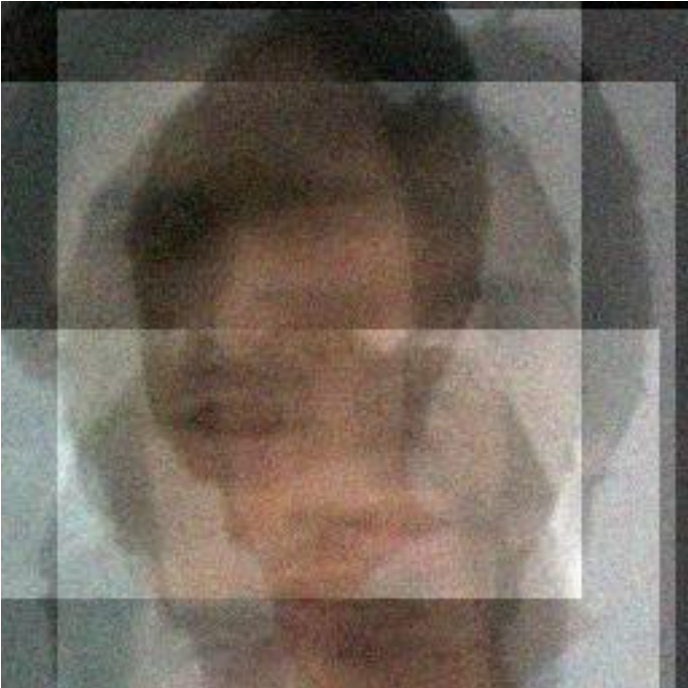}
	\end{minipage}
	
	\begin{minipage}{0.3\linewidth}
		\centering
		\includegraphics[width=0.9\linewidth]{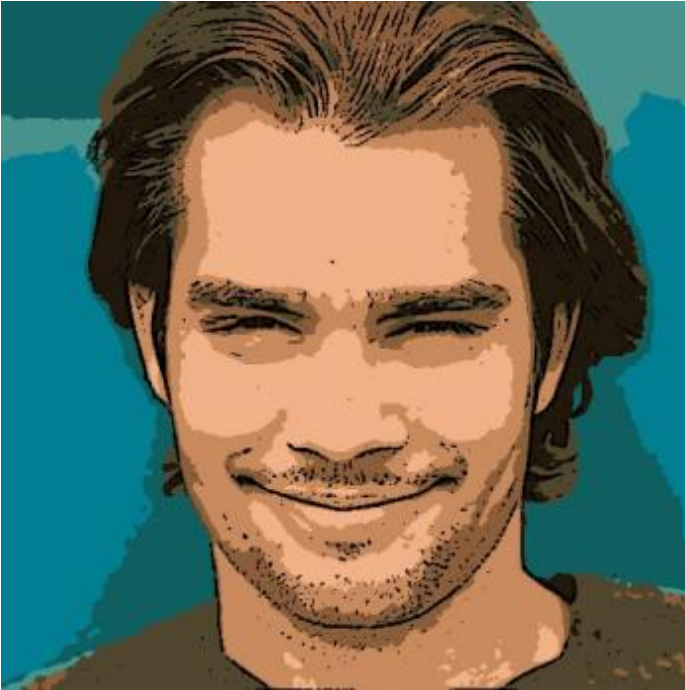}
	\end{minipage}
	\begin{minipage}{0.3\linewidth}
		\centering
		\includegraphics[width=0.9\linewidth]{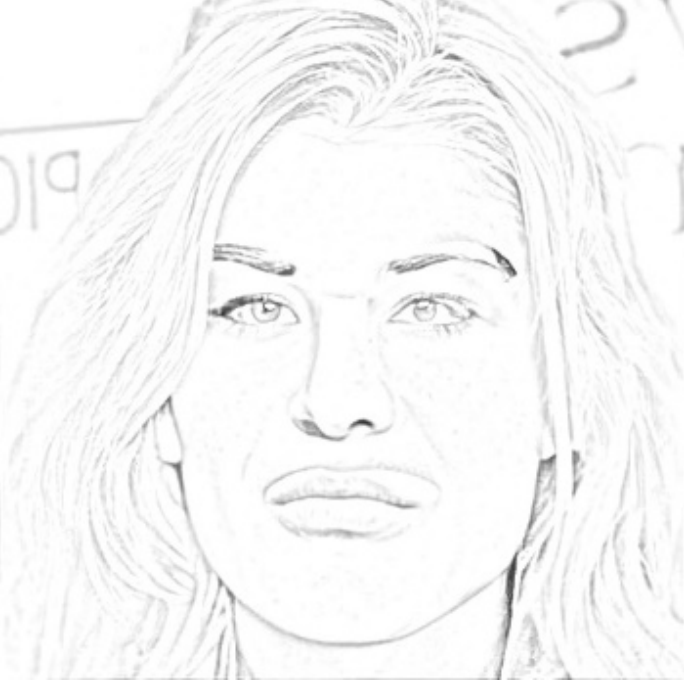}
	\end{minipage}
    \begin{minipage}{0.3\linewidth}
		\centering
		\includegraphics[width=0.9\linewidth]{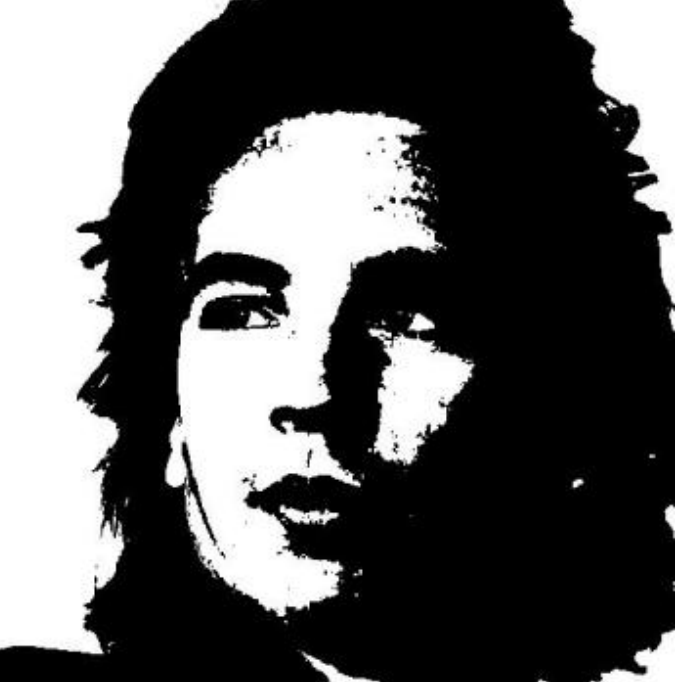}
	\end{minipage}
\caption{Geneerated training negative samples.}
\end{figure}

\subsection{Features representation}
Feature extraction is performed using the Swin Transformer V2-B model, which utilizes a self-attention mechanism to capture detailed features and contextual information globally to improve the accuracy of deepfake detection. Swin Transformer constructs hierarchical representations by starting with small-sized patches and gradually merging neighboring patches in deeper Transformer layers. With the help of these hierarchical feature maps, the Swin Transformer model can be easily utilized for intensive prediction using advanced techniques. Linear computational complexity can be achieved by locally computing selfattention within non-overlapping windows of the segmented image. A key design element of the Swin Transformer is its shifting of window partitions between successive self-attentive layers. The shifted windows bridge the windows of the previous layer, providing a connection between the two and significantly enhancing the modeling capabilities.

\subsection{Dimensionality reduction}
After feature extraction of the dataset using the image backbone network ResNet18, the features are analyzed using clustering methods such as t-SNE, Kmeans, etc., which leads to an understanding of the ways in which the deepfake are carried out in some of the data, including facial region occlusion, region cropping, and generative model generation.

\subsection{Method complexity}
Complexity: the number of model parameters is about 86.89 M, and the inference speed is up to 39.215 FPS on NVIDIA A100.

\noindent Innovation: Diverse training samples are generated for various deepfake methods, including random facial region cutout, local cropping, cartoonization, etc. For the classification prediction results, post-processing is performed by combining Dlib third-party library face keypoint detector with Opencv Haar cascade face detector to correct the confidence level of the network that is difficult to make judgment.

\subsection{Method generalization}
Diverse training samples are generated using various data augment methods for various deepfake methods, which enhances the generalization ability and robustness of the model.

\section{Experiments}

\subsection{Train strategy}
Swin Transformer V2-B model based on ImageNet1k pre-training was used. A two-stage training strategy was used. The model is first fine-tuned using ImageNet1k pre-training weights, and then further fine-tuned on the additional training samples generated as well as the original training set samples. The optimization objective uses a cross-entropy loss function common in classification tasks. The model parameters are updated using the AdamW optimizer with a cosine learning rate decay strategy.

Training and testing details description FOR Multi-dimensional Facial Forgery detection solution are as follows:
\begin{itemize}
	\item AdamW optimizer parameters
    \begin{itemize}
        \item eps: 1e-8
        \item betas: (0.9, 0.999)
        \item weight decay: 0.1
    \end{itemize}
	\item Gradient clip: 5.0
    \item Stage One Fine-tuning
    \begin{itemize}
        \item warmup epochs: 5
        \item base lr: 5e-5
        \item warmup lr: 5e-8
        \item min lr: 5e-7
    \end{itemize}
    \item Stage Two Fine-tuning
    \begin{itemize}
        \item warmup epochs: 5
        \item base lr: 5e-6
        \item warmup lr: 5e-9
        \item min lr: 5e-9
    \end{itemize}
\end{itemize}

\subsection{Train datasets}
\begin{enumerate}
\item The released MultiFF data set(524K);
\item Random facial region cutout (40K);
\item Localized cropping (10K);
\item Random grayscaling, translating and overlaying (10K);
\item Cartoonization (10K);
\item Sketching (5K);
\item Binarization (5K).
\end{enumerate}

\subsection{Results}
By employing specific data augment techniques, a series of diverse training samples are generated to withstand various deepfake methods, thus effectively improving the generalization performance and robustness of the model.
In order to withstand all kinds of deepfake methods, diverse data augment strategies are adopted to enrich the training dataset, and these strategies significantly improve the generalization ability of the model and its robustness in the face of unknown attacks.

\begin{table}[h]
    \centering
    \begin{tabular}{|p{22em}|p{6em}|}
        \hline
        Method&Score\\
        \hline
        Swin Transformer V2 (Random facial region cutout, Local cropping)&0.9691634400\\
        \hline
        Swin Transformer V2 (Random facial region cutout, Cutout, Local cropping)&0.9678942187\\
        \hline
        Swin Transformer V2 (Random facial region cutout, Local cropping, Random grayscaling, translating and overlaying, Binarization)&0.9673853270\\
        \hline
        Swin Transformer V2 (Random facial region cutout, Local cropping, Random grayscaling, translating and overlaying, Binarization,  Cartoonization , Sketching)&0.9661416867\\
        \hline
        Swin Transformer V2 (Random facial region cutout, Local cropping on fake images)&0.9687322018\\
        \hline
        Swin Transformer V2 (Random facial region cutout, Local cropping, Random grayscaling, translating and overlaying, Binarization,  Cartoonization , Sketching on fake images)&0.9669118136\\
        \hline
        EfficientNetb1 (Knowledge Distillation with EfficientNetb1-Softlabeling, EfficientNetb1-Weight Decay, EfficientNetb1-FocalLoss, Swin Transformer v2 and Data Augment with Cutmix and MixUp)&0.7366554705\\
        \hline
    \end{tabular}
    \caption{Results of the comparison to other approaches}
\end{table}

\subsection{Other details}
The parallel training was performed on 2 NVIDIA A100s. Using the Python language and the Pytorch framework, 63243 MiB of memory was used per graphics card with an input image size of 256 × 256 and a batch size of 256.

\noindent We trained With 2 A100s, each epoch takes about 40 min. 30-40 epochs are needed for the first stage of training, and 10-20 epochs are needed for the second stage to fine-tune on the training set as well as the additional dataset.

\noindent We tested with single A100, it takes about 12 minutes.

\section{Conclusion}
In this challenging and innovative competition, it is not easy for our team to win the 18th place after a month of hard work and unremitting exploration. This competition is not only a comprehensive test of our technical ability, but also a profound sharpening of teamwork, innovative thinking and indomitable spirit in the face of complex problems. The MultiFF dataset, with its high degree of diversity and complexity, provided us with an almost real-world arena, allowing us to deeply appreciate the rapid development of deepfake and their potential threat to social security.

\noindent During the competition, we witnessed the infinite possibilities of the technology, and also deeply realized that, with the continuous progress of artificial intelligence technology, deepfake technology is increasingly approaching the realm of the fake. This is not only a technological breakthrough, but also puts higher requirements on us - how to more accurately and efficiently identify and prevent these deepfake methods, and protect personal privacy and information security has become a major issue we all face.

\noindent Looking ahead, we are looking forward to the new challenge of deepfake. We believe that through continuous technical research and development, interdisciplinary cooperation and exchanges, as well as continuous optimization of algorithms and models, we will be able to further improve the accuracy and efficiency of the detection technology, and contribute to the construction of a safer and more trustworthy digital world. At the same
time, we also call on all sectors of society to pay attention to this issue, strengthen the construction of laws and regulations, raise public awareness of prevention, and work together to meet the challenges posed by deepfake technology. On the road ahead, let’s work hand in hand to create a brilliant future!

\bibliographystyle{alpha}
\bibliography{sample}
Liu, Z., Lin, Y., Cao, Y., Hu, H., Wei, Y., Zhang, Z., Lin, S., \& Guo, B. (2021). Swin Transformer: Hierarchical Vision Transformer using Shifted Windows. 2021 IEEE/CVF International Conference on Computer Vision (ICCV), 9992-10002.

\noindent Tan, M., \& Le, Q.V. (2019). EfficientNet: Rethinking Model Scaling for Convolutional Neural Networks. ArXiv, abs/1905.11946.

\noindent Devries, T., \& Taylor, G.W. (2017). Improved Regularization of Convolutional Neural Networks with Cutout. ArXiv, abs/1708.04552.

\noindent Zhang, H., Cissé, M., Dauphin, Y., \& Lopez-Paz, D. (2017). mixup: Beyond Empirical Risk Minimization. ArXiv, abs/1710.09412.

\end{document}